# 探索各種梯度下降演算法超參數的最佳值

Abel C. H. Chen

Telecommunication Laboratories, Chunghwa Telecom Co., Ltd.

## 摘要

　　近年來各種梯度下降演算法(如：梯度下降、動量梯度下降、自適應梯度、方均根傳播、自適應矩估計等)被應用到許多個深度學習模型的參數最佳化，以提供高正確率或低誤差。這些最佳化演算法可能有許多個超參數值需要設定，例如：學習率、動量係數等。並且超參數值將有可能影響最佳化收斂速度和解的精確度。因此，本研究提出一個分析框架，運用數學模型分析各種梯度下降演算法在各個目標函數中的平均誤差，並且從最小化平均誤差來決定最合適的超參數值。通過分析結果，歸納出超參數值設置原則，作為模型最佳化參考。在實驗結果顯示本研究方法可以提供高效率收斂和低誤差。

*關鍵詞：超參數、最佳化、梯度下降、方均根傳播*

# Exploring the Optimized Value of Each Hyperparameter in Various Gradient Descent Algorithms

**Abstract**

　　In the recent years, various gradient descent algorithms including the methods of gradient descent, gradient descent with momentum, adaptive gradient (AdaGrad), root-mean-square propagation (RMSProp) and adaptive moment estimation (Adam) have been applied to the parameter optimization of several deep learning models with higher accuracies or lower errors. These optimization algorithms may need to set the values of several hyperparameters which include a learning rate, momentum coefficients, etc. Furthermore, the convergence speed and solution accuracy may be influenced by the values of hyperparameters. Therefore, this study proposes an analytical framework to use mathematical models for analyzing the mean error of each objective function based on various gradient descent algorithms. Moreover, the suitable value of each hyperparameter could be determined by minimizing the mean error. The principles of hyperparameter value setting have been generalized based on analysis results for model optimization. The experimental results show that higher efficiency convergences and lower errors can be obtained by the proposed method.

# 1. 前言

近年來，隨著機器學習和深度學習演算法(LeCun, Bengio & Hinton, 2015)的發展，提升許多技術的正確率，包含影像辨識(Wei et al., 2022)、語音和語意辨識(Sisman et al., 2021)等。後續更帶來許多產業化的應用，例如：無人機(Bouguettaya et al., 2022)、語音機器人(Kirmani, 2022)。其中，這些機器學習和深度學習的模型在最佳化的過程中，將會搭配一些最佳化演算法來降低誤差或提升正確率(Soydaner, 2020)，包含有梯度下降演算法(Rumelhart, Hinton & Williams, 1986)、群智能最佳化演算法(Marini & Walczak, 2015)等。然而，在這些最佳化演算法裡面可能有許多超參數需要設置，並且部分超參數(Hyperparameter)的值則可能採用過去研究證明可能較合適的值或通過實驗的方式來決定超參數的值(Breuel, 2015)。

有鑑於近幾年深度學習模型多為採用各種梯度下降演算法進行模型參數最佳化，所以本研究主要著重探索各種梯度下降演算法超參數的最佳值，分別討論梯度下降(Gradient Descent, GD)、動量梯度下降(Gradient Descent with Momentum)、自適應梯度(Adaptive Gradient, AdaGrad)、以及方均根傳播(Root-Mean-Square Propagation, RMSProp)(Sun et al., 2020)的超參數。為了有效從理論基礎出發，論證各種梯度下降演算法超參數的最佳值，本研究提出一個分析框架，可以分析各種梯度下降演算法在各個目標函數中的平均誤差，並且從最小化平均誤差來決定最合適的超參數值。並且深入探討超參數最佳解與其他變量和其他超參數之間的關聯，以及歸納出一些超參數設置原則，可以作為後續發展其他深度學習模型參數最佳化時參考。

本論文共分為五個章節。第 2 節中將說明各種梯度下降演算法，以及比較不同方法間的差異。第 3 節將說明本研究提出的分析框架，並且運用分析框架深入討論各種梯度下降演算法在三個不同的目標函數的超參數最佳值，以及在 3.5 節中歸納本研究在超參數設置的新發現。第 4 節將提供實例驗證本研究的超參數最佳值在各個目標函數的收斂表現。第 5 節總結本研究發現和討論未來研究方向。

# 2. 各種梯度下降演算法

本節將介紹各種梯度下降演算法，包含梯度下降、動量梯度下降、自適應梯度、以及方均根傳播(Sun et al., 2020)的作法及原理。為解釋各種梯度下降演算法，本節假設有一目標函數為 $K(w)$，函數有參數 $w$。在下面章節分別以各種梯度下降演算法說明對參數 $w$ 最佳化的計算方式。

## 2.1 梯度下降方法

梯度下降方法，主要通過學習率和梯度對參數進行修正。例如，第 $i$ 個 Epoch 參數值是 $w_i$，並且計算梯度值是 $\frac{\partial K}{\partial w_i}$。通過超參數學習率 $\eta$，可以計算第 $i+1$ 個 Epoch 參數值是 $w_{i+1} = w_i - \eta \frac{\partial K}{\partial w_i}$，並且可代入目標函數計算目標值，如公式(1)所示。

$$K\left(w_i - \eta \frac{\partial K}{\partial w_i}\right) \tag{1}$$

## 2.2 動量梯度下降方法

動量梯度下降方法，主要通過動量係數、動量、學習率、以及梯度對參數進行修正。例如，第 $i$ 個 Epoch 動量值是 $v_i$，參數值是 $w_i$，並且計算梯度值是 $\frac{\partial K}{\partial w_i}$。通過超參數動量係數 $\alpha$、學習率 $\eta$，可以計算第 $i+1$ 個 Epoch 參數值是 $w_{i+1} = w_i + \alpha v_i - \eta \frac{\partial K}{\partial w_i}$，並且可代入目標函數計算目標值，如公式(2)所示。

$$\begin{aligned} & K(w_i + v_{i+1}), \text{where } v_{i+1} = \alpha v_i - \eta \frac{\partial K}{\partial w_i} \\ & = K\left(w_i + \alpha v_i - \eta \frac{\partial K}{\partial w_i}\right) \end{aligned} \tag{2}$$

## 2.3 自適應梯度方法

自適應梯度方法，主要通過累加過去的梯度平方和來自適應調整學習率，以及參考當下的梯度對參數進行修正。例如，第 $i$ 個 Epoch 參數值是 $w_i$，並且計算梯度值是 $\frac{\partial K}{\partial w_i}$。通過超參數學習率 $\eta$，可以計算第 $i+1$ 個 Epoch 自適應學習率是 $\eta \frac{1}{\sqrt{\varphi_i + \varepsilon}}$，參數值是 $w_{i+1} = w_i - \eta \frac{1}{\sqrt{\varphi_i + \varepsilon}} \frac{\partial K}{\partial w_i}$，並且可代入目標函數計算目標值，如公式(3)所示。其中，常數 $\varepsilon$ 是一極小值，用來避免分母為 0 的情況。

$$K\left(w_i - \eta \frac{1}{\sqrt{\varphi_i + \varepsilon}} \frac{\partial K}{\partial w_i}\right), \text{where } \varphi_i = \sum_{j=1}^{i} \frac{\partial K}{\partial w_j}^2 \tag{3}$$

## 2.4 方均根傳播方法

方均根傳播方法，主要通過梯度平方值係數、加權梯度平方值、學習率、以

及梯度對參數進行修正。其中,將通過累加過去的加權梯度平方值來自適應調整學習率。例如,第 $i$ 個 Epoch 加權梯度平方值是$u_i$,參數值是$w_i$,並且計算梯度值是$\frac{\partial K}{\partial w_i}$。通過超參數梯度平方值係數$\beta$、學習率$\eta$,可以計算第 $i+1$ 個 Epoch 自適應學習率是$\eta \frac{1}{\sqrt{\phi_{i+1}+\varepsilon}}$,參數值是$w_{i+1} = w_i - \eta \frac{1}{\sqrt{\phi_{i+1}+\varepsilon}} \frac{\partial K}{\partial w_i}$,並且可代入目標函數計算目標值,如公式(4)所示。

$$K\left(w_i - \eta \frac{1}{\sqrt{\phi_{i+1}+\varepsilon}} \frac{\partial K}{\partial w_i}\right), \text{where } \phi_{i+1} = \beta u_i + (1-\beta)\frac{\partial K}{\partial w_i}^2 \tag{4}$$

## 2.5 自適應矩估計方法

自適應矩估計(Adaptive Moment Estimation, Adam)方法(Kingma & Ba, 2014),主要結合動量梯度下降方法和方均根傳播方法,同時考量動量係數、動量、學習率、梯度平方值係數、加權梯度平方值、以及梯度。在許多實務應用上顯示自適應矩估計方法可以提供最低的誤差、最高的正確率等高效能。然而,因自適應矩估計方法的理論推導與動量梯度下降方法及方均根傳播方法相同,並受限於論文篇幅,所以本研究將不對自適應矩估計方法展開論述。

## 3. 研究方法

為論證各種梯度下降演算法超參數的最佳值,本研究提出一個分析框架,通過分析框架計算目標函數的平均誤差,再根據最小化平均誤差來推導超參數的最佳值。在 3.1 節中詳細本研究提出的分析框架,並且在 3.2 節和 3.3 節分別以單參數函數和多參數函數來演示分析框架和討論在這些函數裡的超參數最佳值。在 3.4 節以線性迴歸函數來分析超參數的最佳值。最後,在 3.5 節總結在各個函數推導過程的發現。

## 3.1 分析框架

本研究提出的分析框架流程如圖 1 所示。先定義目標函數,再假設參數呈均勻分佈時推導平均誤差。其中,以神經網路模型而言,參數 $w$ 和參數 $b$ 等在初始值時多為採用隨機數,所以如果在實驗多次後將呈現均勻分佈。在前述的基礎上,再根據各種梯度下降演算法推導在進行梯度下降後的平均誤差。最後,再根據此平均誤差函數進行最小化,推導在平均誤差函數最小化時的超參數最佳值。

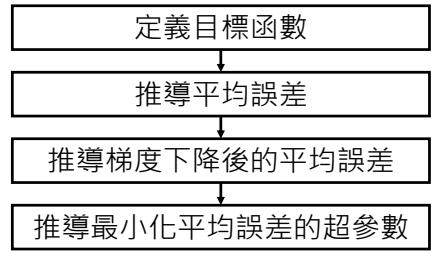

圖 1　分析框架

## 3.2 單參數函數(函數 1)

為說明本研究提出的分析框架，在本節中將先以單參數函數為例進行各種梯度下降演算法的超參數最佳化。

### 3.2.1 函數 1 定義

本研究定義的**函數 1** 是單參數函數$F_1(w)$如公式(5)所示，包含單參數 $w$。在此例中，以$F_1(w)$值作為誤差，並且最佳化目標是最小化誤差(即$F_1(w)$值)。假設參數 $w$ 服從均勻分佈，並且介於 0 到 1 之間，則可以通過公式(6)估計平均誤差。

$$F_1(w) = (w - 0.5)^2 \tag{5}$$

$$\int F_1(w)\, dw = \int (w - 0.5)^2\, dw \tag{6}$$

### 3.2.2 函數 1 的梯度下降方法超參數最佳化

在本節中將採用梯度下降方法對**函數 1** 進行最佳化，並且討論的梯度下降方法超參數的最佳值。

在梯度下降方法中，參數 $w$ 的最佳化主要通過學習率$\eta$和梯度$\frac{\partial F_1}{\partial w}$來修正，即 $w - \eta \frac{\partial F_1}{\partial w}$。因此，將修正後的值代入**函數 1** 中來計算經過梯度下降方法計算後的平均誤差，其中$\frac{\partial F_1}{\partial w} = 2w - 1 = 2(w - 0.5)$，如公式(7)所示。

$$\int F_1\left(w - \eta \frac{\partial F_1}{\partial w}\right) dw = \int \left(w - \eta \frac{\partial F_1}{\partial w} - 0.5\right)^2 dw \tag{7}$$

$$= \frac{\left((1-2\eta)(w-0.5)\right)^3}{3}$$

為最佳化超參數(即學習率$\eta$)和最小化平均誤差，定義函數$G_1(\eta)$，如公式(8)所示。並且對公式(8)的$\eta$微分求一階導函數為 0 時的$\eta$值(即學習率最佳值$\eta^*$)，如公式(9)所示。通過公式(9)可得，在梯度下降方法對**函數 1** 進行最佳化時，學習率最佳值為 0.5。

$$G_1(\eta) = \int F_1\left(w - \eta \frac{\partial F_1}{\partial w}\right) dw \tag{8}$$

$$\begin{aligned}\eta^* &= \underset{0 \leq \eta \leq 1}{\arg\min}\, G_1(\eta) \\ &= \underset{0 \leq \eta \leq 1}{\arg\min}\, \frac{\left((1-2\eta)(w-0.5)\right)^3}{3}\end{aligned} \tag{9}$$

$$\Rightarrow \eta^* = 0.5$$

### 3.2.3 函數 1 的動量梯度下降方法超參數最佳化

在本節中將採用動量梯度下降方法對**函數 1** 進行最佳化，並且討論的動量梯度下降方法超參數(即學習率和動量係數)的最佳值。

在動量梯度下降方法中，參數 $w$ 的最佳化主要通過動量係數$\alpha$、動量$v$、學習率$\eta$、以及梯度$\frac{\partial F_1}{\partial w}$來修正，即$w + \alpha v - \eta \frac{\partial F_1}{\partial w}$。因此，將修正後的值代入**函數 1** 中來計算經過梯度下降方法計算後的平均誤差，如公式(10)所示。

$$\begin{aligned}&\int F_1\left(w + \alpha v - \eta \frac{\partial F_1}{\partial w}\right) dw \\ &= \int \left(w + \alpha v - \eta \frac{\partial F_1}{\partial w} - 0.5\right)^2 dw \\ &= \frac{\left((1-2\eta)(w-0.5) + \alpha v\right)^3}{3}\end{aligned} \tag{10}$$

為最佳化超參數(即學習率$\eta$和動量係數$\alpha$)和最小化平均誤差，定義函數$H_1(\eta, \alpha)$，如公式(11)所示。並且對公式(11)的$\eta$偏微分求一階導函數為 0 時的$\eta$值(即學習率最佳值$\eta^*$)，如公式(12)所示；對公式(11)的$\alpha$偏微分求一階導函數為 0 時的$\alpha$值(即動量係數最佳值$\alpha^*$)，如公式(13)所示。通過公式(12)和(13)可得，在梯

度下降方法對函數 1 進行最佳化時，學習率最佳值為$\frac{\alpha v+(w-0.5)}{2(w-0.5)}$、動量係數最佳值為$\frac{2(\eta-0.5)(w-0.5)}{v}$。

$$H_1(\eta,\alpha) = \int F_1\left(w + \alpha v - \eta \frac{\partial F_1}{\partial w}\right) dw \tag{11}$$

$$\eta^* = \underset{0 \leq \eta \leq 1}{arg\ min}\, H_1(\eta,\alpha)$$

$$= \underset{0 \leq \eta \leq 1}{\arg\min} \frac{\left((1-2\eta)(w-0.5) + \alpha v\right)^3}{3} \tag{12}$$

$$\Rightarrow \eta^* = \frac{\alpha v + (w-0.5)}{2(w-0.5)}$$

$$\alpha^* = \underset{0 \leq \alpha \leq 1}{arg\ min}\, H_1(\eta,\alpha)$$

$$= \underset{0 \leq \alpha \leq 1}{\arg\min} \frac{\left((1-2\eta)(w-0.5) + \alpha v\right)^3}{3} \tag{13}$$

$$\Rightarrow \alpha^* = \frac{2(\eta-0.5)(w-0.5)}{v}$$

### 3.2.4 函數 1 的自適應梯度方法超參數最佳化

在本節中將採用自適應梯度方法對函數 1 進行最佳化，並且討論的自適應梯度方法超參數(即學習率)的最佳值。

在自適應梯度方法中，參數 $w$ 的最佳化主要通過學習率$\eta \frac{1}{\sqrt{\varphi+\varepsilon}}$和梯度$\frac{\partial F_1}{\partial w}$來修正，即$w - \eta \frac{1}{\sqrt{\varphi+\varepsilon}} \frac{\partial F}{\partial w}$。其中，$\varphi$是累加過去的梯度平方和。將修正後的值代入函數 1 中來計算經過梯度下降方法計算後的平均誤差，如公式(14)所示。

$$\int F_1\left(w - \eta \frac{1}{\sqrt{\varphi+\varepsilon}} \frac{\partial F_1}{\partial w}\right) dw$$

$$= \int \left(w - \eta \frac{g_w}{\sqrt{\varphi+\varepsilon}} - 0.5\right)^2 dw \tag{14}$$

$$= \frac{\left(\left(1 - \frac{2\eta}{\sqrt{\varphi+\varepsilon}}\right)(w-0.5)\right)^3}{3}$$

為最佳化超參數(即學習率$\eta$)和最小化平均誤差,定義函數$O_1(\eta)$,如公式(15)所示。並且對公式(15)的$\eta$偏微分求一階導函數為 0 時的$\eta$值(即學習率最佳值$\eta^*$),如公式(16)所示。通過公式(16)可得,在梯度下降方法對**函數 1** 進行最佳化時,學習率最佳值為$\frac{\sqrt{\varphi+\varepsilon}}{2}$。

$$O_1(\eta) = \int F_1\left(w - \eta \frac{1}{\sqrt{\varphi+\varepsilon}} \frac{\partial F_1}{\partial w}\right) dw \tag{15}$$

$$\begin{aligned}\eta^* &= \underset{0\leq\eta\leq 1}{arg\ min}\ O_1(\eta) \\ &= \underset{0\leq\eta\leq 1}{arg\ min} \frac{\left(\left(1-\frac{2\eta}{\sqrt{\varphi+\varepsilon}}\right)(w-0.5)\right)^3}{3} \\ \Rightarrow \eta^* &= \frac{\sqrt{\varphi+\varepsilon}}{2}\end{aligned} \tag{16}$$

### 3.2.5 函數 1 的方均根傳播方法超參數最佳化

在本節中將採用方均根傳播方法對**函數 1** 進行最佳化,並且討論的方均根傳播方法超參數(即學習率和梯度平方值係數)的最佳值。

在方均根傳播方法中,參數 $w$ 的最佳化主要通過梯度平方值係數$\beta$、加權梯度平方值$u$、學習率$\eta \frac{1}{\sqrt{\phi+\varepsilon}}$、以及梯度$\frac{\partial F_1}{\partial w}$來修正,即$w - \eta \frac{1}{\sqrt{\phi+\varepsilon}} \frac{\partial F_1}{\partial w}$。其中,$\phi$是累加過去的加權梯度平方和,表示為$\phi = \beta u + (1-\beta)\frac{\partial F_1}{\partial w}^2$。將修正後的值代入**函數 1** 中來計算經過梯度下降方法計算後的平均誤差,如公式(17)所示。

$$\begin{aligned}&\int F_1\left(w - \eta \frac{1}{\sqrt{\phi+\varepsilon}} \frac{\partial F_1}{\partial w}\right) dw \\ &= \int \left(w - \eta \frac{1}{\sqrt{\phi+\varepsilon}} \frac{\partial F_1}{\partial w} - 0.5\right)^2 dw \\ &= \frac{\left(\left(1-\frac{2\eta}{\sqrt{\phi+\varepsilon}}\right)(w-0.5)\right)^3}{3}\end{aligned} \tag{17}$$

為最佳化超參數(即學習率$\eta$和梯度平方值係數$\beta$)和最小化平均誤差,定義函

數$Q_1(\eta, \beta)$,如公式(18)所示。並且對公式(18)的$\eta$偏微分求一階導函數為 0 時的$\eta$值(即學習率最佳值$\eta^*$),如公式(19)所示;對公式(18)的$\beta$偏微分求一階導函數為 0 時的$\beta$值(即動量係數最佳值$\beta^*$),如公式(20)所示。通過公式(19)和(20)可得,在梯度下降方法對函數 1 進行最佳化時,學習率最佳值為$\frac{\sqrt{\phi+\varepsilon}}{2}$、梯度平方值係數最佳值為$\frac{4\eta^2 - \frac{\partial F_1}{\partial w}^2 - \varepsilon}{u - \frac{\partial F_1}{\partial w}^2}$。

$$Q_1(\eta, \beta) = \int F_1\left(w - \eta \frac{1}{\sqrt{\phi+\varepsilon}} \frac{\partial F_1}{\partial w}\right) dw \tag{18}$$

$$\eta^* = \underset{0 \leq \eta \leq 1}{arg\ min}\ Q_1(\eta, \beta)$$

$$= \underset{0 \leq \eta \leq 1}{arg\ min} \frac{\left(\left(1 - \frac{2\eta}{\sqrt{\phi+\varepsilon}}\right)(w - 0.5)\right)^3}{3} \tag{19}$$

$$\Rightarrow \eta^* = \frac{\sqrt{\phi+\varepsilon}}{2}$$

$$\beta^* = \underset{0 \leq \beta \leq 1}{arg\ min}\ Q_1(\eta, \beta)$$

$$= \underset{0 \leq \beta \leq 1}{arg\ min} \frac{\left(\left(1 - \frac{2\eta}{\sqrt{\phi+\varepsilon}}\right)(w - 0.5)\right)^3}{3} \tag{20}$$

$$\Rightarrow \beta^* = \frac{4\eta^2 - \frac{\partial F_1}{\partial w}^2 - \varepsilon}{u - \frac{\partial F_1}{\partial w}^2}$$

## 3.3 多參數函數(函數 2)

本節中以多參數函數為例進行各種梯度下降演算法的超參數最佳化。

### 3.3.1 函數 2 定義

本研究定義的函數 2 是多參數函數$F_2(w, b)$如公式(21)所示,包含參數 $w$ 和參數 $b$。在此例中,以$F_2(w, b)$值作為誤差,並且最佳化目標是最小化誤差(即$F_2(w, b)$值)。假設參數 $w$ 和參數 $b$ 服從均勻分佈,並且介於 0 到 1 之間,則可以

通過公式(22)估計平均誤差。

$$F_2(w, b) = (w + b)^2 \tag{21}$$

$$\iint F_2(w, b)\, dw\, db = \iint (w + b)^2\, dw\, db \tag{22}$$

### 3.3.2 函數 2 的梯度下降方法超參數最佳化

在本節中將採用梯度下降方法對**函數 2** 進行最佳化，並且討論的梯度下降方法超參數的最佳值。

在梯度下降方法中，參數 $w$ 的最佳化主要通過學習率 $\eta$ 和梯度 $\frac{\partial F_2}{\partial w}$ 及 $\frac{\partial F_2}{\partial b}$ 來修正，即 $w - \eta \frac{\partial F_2}{\partial w}$ 及 $b - \eta \frac{\partial F_2}{\partial b}$。因此，將修正後的值代入**函數 2** 中來計算經過梯度下降方法計算後的平均誤差，其中 $\frac{\partial F_2}{\partial w} = \frac{\partial F_2}{\partial b} = 2w + 2b = 2(w + b)$，如公式(23)所示。

$$\iint F_2\left(w - \eta \frac{\partial F_2}{\partial w}, b - \eta \frac{\partial F_2}{\partial b}\right) dw\, db$$
$$= \iint \left(w - \eta \frac{\partial F_2}{\partial w} + b - \eta \frac{\partial F_2}{\partial b}\right)^2 dw\, db \tag{23}$$
$$= \frac{((1 - 4\eta)(w + b))^4}{12}$$

為最佳化超參數(即學習率 $\eta$)和最小化平均誤差，定義函數 $G_2(\eta)$，如公式(24)所示。並且對公式(24)的 $\eta$ 微分求一階導函數為 0 時的 $\eta$ 值(即學習率最佳值 $\eta^*$)，如公式(25)所示。通過公式(25)可得，在梯度下降方法對**函數 2** 進行最佳化時，學習率最佳值為 0.25。

$$G_2(\eta) = \iint F_2\left(w - \eta \frac{\partial F_2}{\partial w}, b - \eta \frac{\partial F_2}{\partial b}\right) dw\, db \tag{24}$$

$$\eta^* = \underset{0 \leq \eta \leq 1}{\arg\min}\, G_2(\eta) = \underset{0 \leq \eta \leq 1}{\arg\min}\, \frac{((1 - 4\eta)(w + b))^4}{12} \tag{25}$$

$\Rightarrow \eta^* = 0.25$

### 3.3.3 函數 2 的動量梯度下降方法超參數最佳化

在本節中將採用動量梯度下降方法對**函數 2** 進行最佳化，並且討論的動量梯度下降方法超參數(即學習率和動量係數)的最佳值。

在動量梯度下降方法中，參數 $w$ 的最佳化主要通過動量係數$\alpha$、動量$v_w$及$v_b$、學習率$\eta$、以及梯度$\frac{\partial F_2}{\partial w}$及$\frac{\partial F_2}{\partial b}$來修正，即$w + \alpha v_w - \eta \frac{\partial F_2}{\partial w}$及$b + \alpha v_b - \eta \frac{\partial F_2}{\partial b}$。因此，將修正後的值代入**函數 2** 中來計算經過梯度下降方法計算後的平均誤差，如公式(26)所示。

$$\int \int F_2\left(w + \alpha v_w - \eta \frac{\partial F_2}{\partial w}, b + \alpha v_b - \eta \frac{\partial F_2}{\partial b}\right) dw\, db$$
$$= \int \int \left(w + \alpha v_w - \eta \frac{\partial F_2}{\partial w} + b + \alpha v_b - \eta \frac{\partial F_2}{\partial b}\right)^2 dw\, db \qquad (26)$$
$$= \frac{\left((w+b)(1-4\eta) + \alpha(v_w + v_b)\right)^4}{12}$$

為最佳化超參數(即學習率$\eta$和動量係數$\alpha$)和最小化平均誤差，定義函數 $H_2(\eta, \alpha)$，如公式(27)所示。並且對公式(27)的$\eta$偏微分求一階導函數為 0 時的$\eta$值(即學習率最佳值$\eta^*$)，如公式(28)所示；對公式(27)的$\eta$偏微分求一階導函數為 0 時的$\alpha$值(即動量係數最佳值$\alpha$)，如公式(29)所示。通過公式(28)和(29)可得，在梯度下降方法對**函數 2** 進行最佳化時，學習率最佳值為$\frac{\alpha(v_w+v_b)+(w+b)}{4(w+b)}$、動量係數最佳值為$\frac{(4\eta-1)(w+b)}{(v_w+v_b)}$。

$$H_2(\eta, \alpha) = \int \int F_2\left(w + \alpha v_w - \eta \frac{\partial F_2}{\partial w}, b + \alpha v_b - \eta \frac{\partial F_2}{\partial b}\right) dw\, db \qquad (27)$$

$$\eta^* = \underset{0 \leq \eta \leq 1}{arg\ min}\, H_2(\eta, \alpha)$$
$$= \underset{0 \leq \eta \leq 1}{arg\ min} \frac{\left((w+b)(1-4\eta) + \alpha(v_w + v_b)\right)^4}{12} \qquad (28)$$
$$\Rightarrow \eta^* = \frac{\alpha(v_w + v_b) + (w+b)}{4(w+b)}$$

$$\alpha^* = \underset{0 \leq \alpha \leq 1}{arg\ min}\, H_2(\eta, \alpha) \qquad (29)$$

$$= \underset{0\leq\alpha\leq 1}{\arg\min} \frac{\left((w+b)(1-4\eta) + \alpha(v_w+v_b)\right)^4}{12}$$

$$\Rightarrow \alpha^* = \frac{(4\eta-1)(w+b)}{(v_w+v_b)}$$

### 3.3.4 函數 2 的自適應梯度方法超參數最佳化

在本節中將採用自適應梯度方法對函數 2 進行最佳化，並且討論的自適應梯度方法超參數(即學習率)的最佳值。

在自適應梯度方法中，參數 $w$ 及 $b$ 的最佳化主要通過學習率 $\eta\frac{1}{\sqrt{\varphi_w+\varepsilon}}$ 及 $\eta\frac{1}{\sqrt{\varphi_b+\varepsilon}}$ 和梯度 $\frac{\partial F_2}{\partial w}$ 及 $\frac{\partial F_2}{\partial b}$ 來修正，即 $w - \eta\frac{1}{\sqrt{\varphi_w+\varepsilon}}\frac{\partial F_2}{\partial w}$ 及 $b - \eta\frac{1}{\sqrt{\varphi_b+\varepsilon}}\frac{\partial F_2}{\partial b}$。其中，$\varphi_w$ 及 $\varphi_b$ 分別是累加 $w$ 及 $b$ 過去的梯度平方和。將修正後的值代入函數 2 中來計算經過梯度下降方法計算後的平均誤差，如公式(30)所示。

$$\iint F_2\left(w - \eta\frac{1}{\sqrt{\varphi_w+\varepsilon}}\frac{\partial F_2}{\partial w}, b - \eta\frac{1}{\sqrt{\varphi_b+\varepsilon}}\frac{\partial F_2}{\partial b}\right)dwdb$$

$$= \iint \left(w - \eta\frac{1}{\sqrt{\varphi_w+\varepsilon}}\frac{\partial F_2}{\partial w} + b - \eta\frac{1}{\sqrt{\varphi_b+\varepsilon}}\frac{\partial F_2}{\partial b}\right)^2 dwdb \quad (30)$$

$$= \frac{\left((w+b)\left(1 - \frac{2\eta}{\sqrt{\varphi_w+\varepsilon}} - \frac{2\eta}{\sqrt{\varphi_b+\varepsilon}}\right)\right)^4}{12}$$

為最佳化超參數(即學習率$\eta$)和最小化平均誤差，定義函數 $O_2(\eta)$，如公式(31)所示。並且對公式(31)的 $\eta$ 偏微分求一階導函數為 0 時的 $\eta$ 值(即學習率最佳值$\eta^*$)，如公式(32)所示。通過公式(32)可得，在梯度下降方法對函數 2 進行最佳化時，學習率最佳值為 $\frac{\sqrt{\varphi_w+\varepsilon}\sqrt{\varphi_b+\varepsilon}}{2(\sqrt{\varphi_w+\varepsilon}+\sqrt{\varphi_b+\varepsilon})}$。

$$O_2(\eta) = \iint F_2\left(w - \eta\frac{1}{\sqrt{\varphi_w+\varepsilon}}\frac{\partial F_2}{\partial w}, b - \eta\frac{1}{\sqrt{\varphi_b+\varepsilon}}\frac{\partial F_2}{\partial b}\right)dwdb \quad (31)$$

$$\eta^* = \underset{0\leq\eta\leq 1}{\arg\min} O_2(\eta) \quad (32)$$

$$= \underset{0\leq\eta\leq1}{\arg\min} \frac{\left((w+b)\left(1 - \frac{2\eta}{\sqrt{\varphi_w + \varepsilon}} - \frac{2\eta}{\sqrt{\varphi_b + \varepsilon}}\right)\right)^4}{12}$$

$$\Rightarrow \eta^* = \frac{\sqrt{\varphi_w + \varepsilon}\sqrt{\varphi_b + \varepsilon}}{2(\sqrt{\varphi_w + \varepsilon} + \sqrt{\varphi_b + \varepsilon})}$$

### 3.3.5 函數 2 的方均根傳播方法超參數最佳化

在本節中將採用方均根傳播方法對**函數 2** 進行最佳化，並且討論的方均根傳播方法超參數(即學習率和梯度平方值係數)的最佳值。

在方均根傳播方法中，參數 $w$ 及 $b$ 的最佳化主要通過梯度平方值係數 $\beta$、加權梯度平方值 $u_w$ 及 $u_b$、學習率 $\eta\frac{1}{\sqrt{\phi_w+\varepsilon}}$ 及 $\eta\frac{1}{\sqrt{\phi_b+\varepsilon}}$、以及梯度 $\frac{\partial F_2}{\partial w}$ 及 $\frac{\partial F_2}{\partial b}$ 來修正，即 $w - \eta\frac{1}{\sqrt{\phi_w+\varepsilon}}\frac{\partial F_2}{\partial w}$ 及 $b - \eta\frac{1}{\sqrt{\phi_b+\varepsilon}}\frac{\partial F_2}{\partial b}$。其中，$\phi$ 是累加 $w$ 及 $b$ 過去的加權梯度平方和，表示為 $\phi_w = \beta u_w + (1-\beta)\frac{\partial F_2}{\partial w}^2$ 及 $\phi_b = \beta u_b + (1-\beta)\frac{\partial F_2}{\partial w}^2$。將修正後的值代入**函數 2** 中來計算經過梯度下降方法計算後的平均誤差，如公式(33)所示。

$$\iint F_2\left(w - \eta\frac{1}{\sqrt{\phi_w+\varepsilon}}\frac{\partial F_2}{\partial w}, b - \eta\frac{1}{\sqrt{\phi_b+\varepsilon}}\frac{\partial F_2}{\partial b}\right)dwdb$$

$$= \iint \left(w - \eta\frac{1}{\sqrt{\phi_w+\varepsilon}}\frac{\partial F_2}{\partial w} + b - \eta\frac{1}{\sqrt{\phi_b+\varepsilon}}\frac{\partial F_2}{\partial b}\right)^2 dwdb \quad (33)$$

$$= \frac{\left((w+b)\left(1 - \frac{2\eta}{\sqrt{\phi_w+\varepsilon}} - \frac{2\eta}{\sqrt{\phi_b+\varepsilon}}\right)\right)^4}{12}$$

為最佳化超參數(即學習率 $\eta$ 和梯度平方值係數 $\beta$)和最小化平均誤差，定義函數 $Q_2(\eta,\beta)$，如公式(34)所示。並且對公式(34)的 $\eta$ 偏微分求一階導函數為 0 時的 $\eta$ 值(即學習率最佳值 $\eta^*$)，如公式(35)所示；對公式(34)的 $\beta$ 偏微分求一階導函數為 0 時的 $\beta$ 值(即動量係數最佳值 $\beta^*$)，並為簡化計算在此例中假設 $u = u_w = u_b$ 及 $g = \frac{\partial F_2}{\partial w} = \frac{\partial F_2}{\partial b}$，如公式(36)所示。通過公式(35)和(36)可得，在梯度下降方法對**函數 2** 進行最佳化時，學習率最佳值為 $\frac{\sqrt{\phi_w+\varepsilon}\sqrt{\phi_b+\varepsilon}}{2(\sqrt{\phi_w+\varepsilon}+\sqrt{\phi_b+\varepsilon})}$、梯度平方值係數最佳值為 $\frac{16\eta^2-g^2-\varepsilon}{(u-g^2)}$。

$$Q_2(\eta,\beta) = \int\int F_2\left(w - \eta \frac{1}{\sqrt{\phi_w + \varepsilon}}\frac{\partial F_2}{\partial w}, b - \eta \frac{1}{\sqrt{\phi_b + \varepsilon}}\frac{\partial F_2}{\partial b}\right)dwdb \tag{34}$$

$$\eta^* = \underset{0\leq\eta\leq 1}{arg\,min}\, Q_2(\eta,\beta)$$

$$= \underset{0\leq\eta\leq 1}{arg\,min} \frac{\left((w+b)\left(1 - \frac{2\eta}{\sqrt{\phi_w + \varepsilon}} - \frac{2\eta}{\sqrt{\phi_b + \varepsilon}}\right)\right)^4}{12} \tag{35}$$

$$\Rightarrow \eta^* = \frac{\sqrt{\phi_w + \varepsilon}\sqrt{\phi_b + \varepsilon}}{2(\sqrt{\phi_w + \varepsilon} + \sqrt{\phi_b + \varepsilon})}$$

$$\beta^* = \underset{0\leq\beta\leq 1}{arg\,min}\, Q_2(\eta,\beta)$$

$$= \underset{0\leq\beta\leq 1}{arg\,min} \frac{\left((w+b)\left(1 - \frac{2\eta}{\sqrt{\phi_w + \varepsilon}} - \frac{2\eta}{\sqrt{\phi_b + \varepsilon}}\right)\right)^4}{12} \tag{36}$$

$$\Rightarrow \beta^* = \frac{16\eta^2 - g^2 - \varepsilon}{(u - g^2)}$$

## 3.4 線性迴歸函數(函數 3)

本節中以線性迴歸函數為例進行各種梯度下降演算法的超參數最佳化，未來研究者可參考此案例擴展成神經網路模型。

### 3.4.1 函數 3 定義

本研究定義的**函數 3** 是線性迴歸函數$F_3(w,b)$如公式(37)所示，包含參數 $w$、參數 $b$、變量 $x$、變量 $y$。在線性迴歸應用，以變量 $x$ 作為輸入、$wx + b$作為輸出的估計值、變量 $y$ 作為輸出的真值，並且假設以平方誤差作為損失(或稱為誤差)。在此例中，以$F_3(w,b)$值作為誤差，並且最佳化目標是最小化誤差(即$F_3(w,b)$值)。假設參數 $w$ 和參數 $b$ 服從均勻分佈，並且介於 0 到 1 之間，則可以通過公式(38)估計平均誤差。

$$F_3(w,b) = (wx + b - y)^2 \tag{37}$$

$$\iint F_3(w, b)\, dw\, db$$
$$= \iint (wx + b - y)^2\, dw\, db \tag{38}$$

### 3.4.2 函數 3 的梯度下降方法超參數最佳化

在梯度下降方法中，參數 $w$ 的最佳化主要通過學習率$\eta$和梯度$\frac{\partial F_3}{\partial w}$及$\frac{\partial F_3}{\partial b}$來修正，即$w - \eta \frac{\partial F_3}{\partial w}$及$b - \eta \frac{\partial F_3}{\partial b}$。因此，將修正後的值代入**函數 3** 中來計算經過梯度下降方法計算後的平均誤差，令$\delta = wx + b - y$，則$\frac{\partial F_3}{\partial w} = x\delta$及$\frac{\partial F_3}{\partial b} = \delta$，如公式(39)所示。

$$\iint F_3\left(w - \eta \frac{\partial F_3}{\partial w}, b - \eta \frac{\partial F_3}{\partial b}\right) dw\, db$$
$$= \iint \left(\left(w - \eta \frac{\partial F_3}{\partial w}\right)x + \left(b - \eta \frac{\partial F_3}{\partial b}\right) - y\right)^2 dw\, db \tag{39}$$
$$= \frac{\left(((wx+b)-y) - \eta\left(\frac{\partial F_3}{\partial w}x + \frac{\partial F_3}{\partial b}\right)\right)^4}{12x}$$

為最佳化超參數(即學習率$\eta$)和最小化平均誤差，定義函數$G_3(\eta)$，如公式(40)所示。並且對公式(40)的$\eta$微分求一階導函數為 0 時的$\eta$值(即學習率最佳值$\eta^*$)，如公式(25)所示。通過公式(41)可得，在梯度下降方法對**函數 3** 進行最佳化時，學習率最佳值為$\frac{1}{x^2+1}$。

$$G_3(\eta) = \iint F_3\left(w - \eta \frac{\partial F_3}{\partial w}, b - \eta \frac{\partial F_3}{\partial b}\right) dw\, db \tag{40}$$

$$\eta^* = \underset{0 \leq \eta \leq 1}{\arg\min}\, G_3(\eta)$$
$$= \underset{0 \leq \eta \leq 1}{\arg\min} \frac{\left(((wx+b)-y) - \eta\left(\frac{\partial F_3}{\partial w}x + \frac{\partial F_3}{\partial b}\right)\right)^4}{12x} \tag{41}$$
$$\Rightarrow \eta^* = \frac{(wx+b)-y}{\frac{\partial F_3}{\partial w}x + \frac{\partial F_3}{\partial b}} = \frac{\delta}{x^2\delta + \delta} = \frac{1}{x^2+1}$$

### 3.4.3 函數 3 的動量梯度下降方法超參數最佳化

在本節中將採用動量梯度下降方法對**函數 3**進行最佳化，並且討論的動量梯度下降方法超參數(即學習率和動量係數)的最佳值。

在動量梯度下降方法中，參數 $w$ 的最佳化主要通過動量係數 $\alpha$、動量 $v_w$ 及 $v_b$、學習率 $\eta$、以及梯度 $\frac{\partial F_3}{\partial w}$ 及 $\frac{\partial F_3}{\partial b}$ 來修正，即 $w + \alpha v_w - \eta \frac{\partial F_3}{\partial w}$ 及 $b + \alpha v_b - \eta \frac{\partial F_3}{\partial b}$。因此，將修正後的值代入**函數 3**中來計算經過梯度下降方法計算後的平均誤差，如公式(42)所示。

$$\int\int F_3\left(w + \alpha v_w - \eta \frac{\partial F_3}{\partial w}, b + \alpha v_b - \eta \frac{\partial F_3}{\partial b}\right) dw\, db$$

$$= \int\int\left(\left(w + \alpha v_w - \eta \frac{\partial F_3}{\partial w}\right)x + \left(b + \alpha v_b - \eta \frac{\partial F_3}{\partial b}\right) - y\right)^2 dw\, db \quad (42)$$

$$= \frac{\left(((wx+b)-y) + \alpha(v_w x + v_b) - \eta\left(\frac{\partial F_3}{\partial w}x + \frac{\partial F_3}{\partial b}\right)\right)^4}{12x}$$

為最佳化超參數(即學習率 $\eta$ 和動量係數 $\alpha$)和最小化平均誤差，定義函數 $H_3(\eta, \alpha)$，如公式(42)所示。並且對公式(42)的 $\eta$ 偏微分求一階導函數為 0 時的 $\eta$ 值 (即學習率最佳值 $\eta^*$)，如公式(43)所示；對公式(42)的 $\eta$ 偏微分求一階導函數為 0 時的 $\alpha$ 值(即動量係數最佳值 $\alpha$)，如公式(44)所示。通過公式(43)和(44)可得，在梯度下降方法對**函數 3**進行最佳化時，學習率最佳值為 $\frac{1}{x^2+1} + \frac{\alpha(v_w x + v_b)}{\delta(x^2+1)}$、動量係數最佳值為 $\frac{\delta(\eta x^2 + \eta - 1)}{v_w x + v_b}$。

$$H_3(\eta, \alpha) = \int\int F_3\left(w + \alpha v_w - \eta \frac{\partial F_3}{\partial w}, b + \alpha v_b - \eta \frac{\partial F_3}{\partial b}\right) dw\, db \quad (42)$$

$$\eta^* = \underset{0 \leq \eta \leq 1}{\arg\min}\, H_3(\eta, \alpha)$$

$$= \underset{0 \leq \eta \leq 1}{\arg\min} \frac{\left(((wx+b)-y) + \alpha(v_w x + v_b) - \eta\left(\frac{\partial F_3}{\partial w}x + \frac{\partial F_3}{\partial b}\right)\right)^4}{12x} \quad (43)$$

$$\Rightarrow \eta^* = \frac{\delta + \alpha(v_w x + v_b)}{x^2\delta + \delta} = \frac{1}{x^2+1} + \frac{\alpha(v_w x + v_b)}{\delta(x^2+1)}$$

$$\alpha^* = \underset{0 \leq \alpha \leq 1}{\arg\min}\, H_3(\eta, \alpha) \quad (44)$$

$$= \underset{0\leq\alpha\leq1}{\arg\min} \frac{\left(((wx+b)-y)+\alpha(v_w x+v_b)-\eta\left(\frac{\partial F_3}{\partial w}x+\frac{\partial F_3}{\partial b}\right)\right)^4}{12x}$$

$$\Rightarrow \alpha^* = \frac{\delta(\eta x^2+\eta-1)}{v_w x+v_b}$$

### 3.4.4 函數 3 的自適應梯度方法超參數最佳化

在本節中將採用自適應梯度方法對函數 3 進行最佳化，並且討論的自適應梯度方法超參數(即學習率)的最佳值。

在自適應梯度方法中，參數 $w$ 及 $b$ 的最佳化主要通過學習率 $\eta\frac{1}{\sqrt{\varphi_w+\varepsilon}}$ 及 $\eta\frac{1}{\sqrt{\varphi_b+\varepsilon}}$ 和梯度 $\frac{\partial F_3}{\partial w}$ 及 $\frac{\partial F_3}{\partial b}$ 來修正，即 $w-\eta\frac{1}{\sqrt{\varphi_w+\varepsilon}}\frac{\partial F_3}{\partial w}$ 及 $b-\eta\frac{1}{\sqrt{\varphi_b+\varepsilon}}\frac{\partial F_3}{\partial b}$。其中，$\varphi_w$ 及 $\varphi_b$ 分別是累加 $w$ 及 $b$ 過去的梯度平方和。將修正後的值代入函數 3 中來計算經過梯度下降方法計算後的平均誤差，如公式(45)所示。

$$\iint F_3\left(w-\eta\frac{1}{\sqrt{\varphi_w+\varepsilon}}\frac{\partial F_3}{\partial w}, b-\eta\frac{1}{\sqrt{\varphi_b+\varepsilon}}\frac{\partial F_3}{\partial b}\right)dwdb$$

$$=\iint\left(\left(w-\eta\frac{1}{\sqrt{\varphi_w+\varepsilon}}\frac{\partial F_3}{\partial w}\right)x+\left(b-\eta\frac{1}{\sqrt{\varphi_b+\varepsilon}}\frac{\partial F_3}{\partial b}\right)-y\right)^2 dwdb \quad (45)$$

$$=\frac{\left(((wx+b)-y)-\eta\left(\frac{1}{\sqrt{\varphi_w+\varepsilon}}\frac{\partial F_3}{\partial w}x+\frac{1}{\sqrt{\varphi_b+\varepsilon}}\frac{\partial F_3}{\partial b}\right)\right)^4}{12x}$$

為最佳化超參數(即學習率 $\eta$)和最小化平均誤差，定義函數 $O_3(\eta)$，如公式(46)所示。並且對公式(46)的 $\eta$ 偏微分求一階導函數為 0 時的 $\eta$ 值(即學習率最佳值 $\eta^*$)，如公式(47)所示。通過公式(47)可得，在梯度下降方法對函數 3 進行最佳化時，學習率最佳值為 $\frac{\sqrt{\varphi_w+\varepsilon}\sqrt{\varphi_b+\varepsilon}}{x^2\sqrt{\varphi_b+\varepsilon}+\sqrt{\varphi_w+\varepsilon}}$。

$$O_3(\eta)=\iint F_3\left(w-\eta\frac{1}{\sqrt{\varphi_w+\varepsilon}}\frac{\partial F_3}{\partial w}, b-\eta\frac{1}{\sqrt{\varphi_b+\varepsilon}}\frac{\partial F_3}{\partial b}\right)dwdb \quad (46)$$

$$\eta^*=\underset{0\leq\eta\leq1}{\arg\min}\, O_3(\eta) \quad (47)$$

$$= \underset{0 \le \eta \le 1}{\arg\min} \frac{\left(((wx+b)-y) - \eta\left(\frac{1}{\sqrt{\varphi_w+\varepsilon}}\frac{\partial F_3}{\partial w}x + \frac{1}{\sqrt{\varphi_b+\varepsilon}}\frac{\partial F_3}{\partial b}\right)\right)^4}{12x}$$

$$\Rightarrow \eta^* = \frac{1}{\left(\frac{x^2}{\sqrt{\varphi_w+\varepsilon}} + \frac{1}{\sqrt{\varphi_b+\varepsilon}}\right)} = \frac{\sqrt{\varphi_w+\varepsilon}\sqrt{\varphi_b+\varepsilon}}{x^2\sqrt{\varphi_b+\varepsilon} + \sqrt{\varphi_w+\varepsilon}}$$

### 3.4.5 函數 3 的方均根傳播方法超參數最佳化

在本節中將採用方均根傳播方法對**函數 3** 進行最佳化，並且討論的方均根傳播方法超參數(即學習率和梯度平方值係數)的最佳值。

在方均根傳播方法中，參數 $w$ 及 $b$ 的最佳化主要通過梯度平方值係數 $\beta$、加權梯度平方值 $u_w$ 及 $u_b$、學習率 $\eta\frac{1}{\sqrt{\phi_w+\varepsilon}}$ 及 $\eta\frac{1}{\sqrt{\phi_b+\varepsilon}}$、以及梯度 $\frac{\partial F_3}{\partial w}$ 及 $\frac{\partial F_3}{\partial b}$ 來修正，即 $w - \eta\frac{1}{\sqrt{\phi_w+\varepsilon}}\frac{\partial F_3}{\partial w}$ 及 $b - \eta\frac{1}{\sqrt{\phi_b+\varepsilon}}\frac{\partial F_3}{\partial b}$。其中，$\phi$ 是累加 $w$ 及 $b$ 過去的加權梯度平方和，表示為 $\phi_w = \beta u_w + (1-\beta)\frac{\partial F_3}{\partial w}^2$ 及 $\phi_b = \beta u_b + (1-\beta)\frac{\partial F_3}{\partial w}^2$。將修正後的值代入**函數 3** 中來計算經過梯度下降方法計算後的平均誤差，如公式(48)所示。

$$\iint F_3\left(w - \eta\frac{1}{\sqrt{\phi_w+\varepsilon}}\frac{\partial F_3}{\partial w}, b - \eta\frac{1}{\sqrt{\phi_b+\varepsilon}}\frac{\partial F_3}{\partial b}\right)dwdb$$

$$= \iint \left(\left(w - \eta\frac{1}{\sqrt{\phi_w+\varepsilon}}\frac{\partial F_3}{\partial w}\right)x + \left(b - \eta\frac{1}{\sqrt{\phi_b+\varepsilon}}\frac{\partial F_3}{\partial b}\right) - y\right)^2 dwdb \quad (48)$$

$$= \frac{\left(((wx+b)-y) - \eta\left(\frac{1}{\sqrt{\phi_w+\varepsilon}}\frac{\partial F_3}{\partial w}x + \frac{1}{\sqrt{\phi_b+\varepsilon}}\frac{\partial F_3}{\partial b}\right)\right)^4}{12x}$$

為最佳化超參數(即學習率 $\eta$ 和梯度平方值係數 $\beta$)和最小化平均誤差，定義函數 $Q_3(\eta,\beta)$，如公式(49)所示。並且對公式(49)的 $\eta$ 偏微分求一階導函數為 0 時的 $\eta$ 值(即學習率最佳值 $\eta^*$)，如公式(50)所示；對公式(49)的 $\beta$ 偏微分求一階導函數為 0 時的 $\beta$ 值(即動量係數最佳值 $\beta^*$)，並為簡化計算在此例中假設 $u = u_w = u_b$ 及 $g = \frac{\partial F_3}{\partial w} = \frac{\partial F_3}{\partial b}$，如公式(51)所示。通過公式(50)和(51)可得，在梯度下降方法對**函數 3** 進行最佳化時，學習率最佳值為 $\frac{\sqrt{\phi_w+\varepsilon}\sqrt{\phi_b+\varepsilon}}{x^2\sqrt{\phi_b+\varepsilon}+\sqrt{\phi_w+\varepsilon}}$、梯度平方值係數最佳值為 $\frac{\eta^2 g^2(x+1)^2 - g^2\delta^2 - \varepsilon\delta^2}{(u-g^2)\delta^2}$。

$$Q_3(\eta,\beta) = \int\int F_3\left(w - \eta\frac{1}{\sqrt{\phi_w+\varepsilon}}\frac{\partial F_3}{\partial w}, b - \eta\frac{1}{\sqrt{\phi_b+\varepsilon}}\frac{\partial F_3}{\partial b}\right)dwdb \tag{49}$$

$$\eta^* = \underset{0\leq\eta\leq 1}{arg\,min}\, Q_3(\eta,\beta)$$

$$= \underset{0\leq\eta\leq 1}{arg\,min}\frac{\left(((wx+b)-y) - \eta\left(\frac{1}{\sqrt{\phi_w+\varepsilon}}\frac{\partial F_3}{\partial w}x + \frac{1}{\sqrt{\phi_b+\varepsilon}}\frac{\partial F_3}{\partial b}\right)\right)^4}{12x} \tag{50}$$

$$\Rightarrow \eta^* = \frac{1}{\left(\frac{x^2}{\sqrt{\phi_w+\varepsilon}} + \frac{1}{\sqrt{\phi_b+\varepsilon}}\right)} = \frac{\sqrt{\phi_w+\varepsilon}\sqrt{\phi_b+\varepsilon}}{x^2\sqrt{\phi_b+\varepsilon} + \sqrt{\phi_w+\varepsilon}}$$

$$\beta^* = \underset{0\leq\beta\leq 1}{arg\,min}\, Q_3(\eta,\beta)$$

$$= \underset{0\leq\beta\leq 1}{arg\,min}\frac{\left(((wx+b)-y) - \eta\left(\frac{1}{\sqrt{\phi_w+\varepsilon}}\frac{\partial F_3}{\partial w}x + \frac{1}{\sqrt{\phi_b+\varepsilon}}\frac{\partial F_3}{\partial b}\right)\right)^4}{12x} \tag{51}$$

$$\Rightarrow \beta^* = \frac{\eta^2 g^2(x+1)^2 - g^2\delta^2 - \varepsilon\delta^2}{(u-g^2)\delta^2}$$

## 3.5 小結與討論

本節對上述各種梯度下降演算法在不同目標函數的超參數最佳值(如表 1 所示)進行深入討論。

由於動量梯度下降、自適應梯度、方均根傳播都是梯度下降的延伸，所以梯度下降的超參數是動量梯度下降、自適應梯度、方均根傳播的超參數組合的特例。從這些公式也可以相互印證本研究推導結果的正確性。

- 當動量梯度下降動量係數$\alpha^* = 0$時，動量梯度下降的學習率與梯度下降的學習率相同。

- 當自適應梯度累加過去的梯度平方和開根號$\sqrt{\varphi_w+\varepsilon} = 1$且$\sqrt{\varphi_b+\varepsilon} = 1$時，自適應梯度的學習率與梯度下降的學習率相同。

- 當方均根傳播累加過去的加權梯度平方和開根號$\sqrt{\phi_w+\varepsilon} = 1$且$\sqrt{\phi_b+\varepsilon} = 1$時，方均根傳播的學習率與梯度下降的學習率相同。

表 1　各種梯度下降演算法在不同目標函數的超參數最佳值

| 方法 | 函數 1 | 函數 2 | 函數 3 |
|---|---|---|---|
| 梯度下降 | 學習率$\eta^*$：<br>0.5 | 學習率$\eta^*$：<br>0.25 | 學習率$\eta^*$：<br>$\dfrac{1}{x^2+1}$ |
| 動量梯度下降 | 學習率$\eta^*$：<br>$\dfrac{\alpha v + (w-0.5)}{2(w-0.5)}$<br>動量係數$\alpha^*$：<br>$\dfrac{2(\eta-0.5)(w-0.5)}{v}$ | 學習率$\eta^*$：<br>$\dfrac{\alpha(v_w+v_b)+(w+b)}{4(w+b)}$<br>動量係數$\alpha^*$：<br>$\dfrac{(4\eta-1)(w+b)}{(v_w+v_b)}$ | 學習率$\eta^*$：<br>$\dfrac{1}{x^2+1}+\dfrac{\alpha(v_w x+v_b)}{\delta(x^2+1)}$<br>動量係數$\alpha^*$：<br>$\dfrac{\delta(\eta x^2+\eta-1)}{v_w x+v_b}$ |
| 自適應梯度 | 學習率$\eta^*$：<br>$\dfrac{\sqrt{\varphi+\varepsilon}}{2}$ | 學習率$\eta^*$：<br>$\dfrac{\sqrt{\varphi_w+\varepsilon}\sqrt{\varphi_b+\varepsilon}}{2(\sqrt{\varphi_w+\varepsilon}+\sqrt{\varphi_b+\varepsilon})}$ | 學習率$\eta^*$：<br>$\dfrac{\sqrt{\varphi_w+\varepsilon}\sqrt{\varphi_b+\varepsilon}}{x^2\sqrt{\varphi_b+\varepsilon}+\sqrt{\varphi_w+\varepsilon}}$ |
| 方均根傳播 | 學習率$\eta^*$：<br>$\dfrac{\sqrt{\phi+\varepsilon}}{2}$<br>梯度平方值係數$\beta^*$：<br>$\dfrac{4\eta^2-\dfrac{\partial F_1}{\partial w}^2-\varepsilon}{u-\dfrac{\partial F_1}{\partial w}^2}$ | 學習率$\eta^*$：<br>$\dfrac{\sqrt{\phi_w+\varepsilon}\sqrt{\phi_b+\varepsilon}}{2(\sqrt{\phi_w+\varepsilon}+\sqrt{\phi_b+\varepsilon})}$<br>梯度平方值係數$\beta^*$：<br>$\dfrac{16\eta^2-g^2-\varepsilon}{(u-g^2)}$ | 學習率$\eta^*$：<br>$\dfrac{\sqrt{\phi_w+\varepsilon}\sqrt{\phi_b+\varepsilon}}{x^2\sqrt{\phi_b+\varepsilon}+\sqrt{\phi_w+\varepsilon}}$<br>梯度平方值係數$\beta^*$：<br>$\dfrac{\eta^2 g^2(x+1)^2-g^2\delta^2-\varepsilon\delta^2}{(u-g^2)\delta^2}$ |

　　由公式推導結果可以歸納出超參數最佳值設置原則，可以作為未來其他模型設置超參數時參考，原則整理如下：

・ 當輸入變量的值越大和輸入變量的數量越多時，則學習率最佳值越小。

・ 當動量值越大時，則動量係數最佳值越小。

・ 當誤差越大、學習率越大、輸入變量的值越大和輸入變量的數量越多時，則動量係數最佳值越大。

・ 當 $\sqrt{\phi_w+\varepsilon}\sqrt{\phi_b+\varepsilon}>\sqrt{\varphi_b+\varepsilon}+\sqrt{\varphi_w+\varepsilon}$ 和 $\sqrt{\phi_w+\varepsilon}\sqrt{\phi_b+\varepsilon}>\sqrt{\phi_b+\varepsilon}+\sqrt{\phi_w+\varepsilon}$時，則學習率的最佳值越大；反之，則學習率最佳值越小。

・ 當學習率越大和輸入變量的數量越多時，則梯度平方值係數最佳值越大。

# 4. 實驗分析與討論

為驗證本研究提出的方法效率和最佳化結果，把本研究方法與預設超參數值的結果分別比較各種梯度下降演算法在不同目標函數的收斂結果。其中，在本節中的超參數初始值和變量設定如下：學習率$\eta = 0.1$、動量係數$\alpha = 0.5$、梯度平方值係數$\beta = 0.5$、以及函數 3 中的輸入變量$w = 0.3$且輸出變量$y = 0.1 \times x + 0.2$。

根據上述超參數初始值和變量設定下，各種梯度下降演算法在不同目標函數的收斂結果如表 2 所示。由實驗結果可以觀察到，本研究提出的方法之超參數在大部分的情況下都可以在第 2 個 Epoch 收斂到零誤差，即只需一次的參數修正就能把目標函數最小化。本研究方法的超參數相較於採用預設超參數或隨機產生的超參數，將可以更快收斂到最小誤差(即最小損失)。

表 2　本研究方法與預設超參數值的收斂結果

| 方法 | 函數 1 | | 函數 2 | | 函數 3 | |
|---|---|---|---|---|---|---|
| | 本研究方法 | 預設超參數 | 本研究方法 | 預設超參數 | 本研究方法 | 預設超參數 |
| 梯度下降 | Epoch：**2** <br> $F_1=0$ | Epoch：63 <br> $F_1=2\times10^{-13}$ | Epoch：**2** <br> $F_2=0$ | Epoch：58 <br> $F_2=2\times10^{-13}$ | Epoch：**2** <br> $F_3=0$ | Epoch：127 <br> $F_3=0$ |
| 動量梯度下降 | Epoch：**2** <br> $F_1=0$ | Epoch：32 <br> $F_1=2\times10^{-13}$ | Epoch：**2** <br> $F_2=0$ | Epoch：26 <br> $F_2=2\times10^{-13}$ | Epoch：**4** <br> $F_3=4\times10^{-14}$ | Epoch：205 <br> $F_3=2\times10^{-13}$ |
| 自適應梯度 | Epoch：**2** <br> $F_1=0$ | Epoch：123 <br> $F_1=2\times10^{-13}$ | Epoch：**2** <br> $F_2=0$ | Epoch：461 <br> $F_2=2\times10^{-13}$ | Epoch：**2** <br> $F_3=0$ | Epoch：311 <br> $F_3=2\times10^{-13}$ |
| 方均根傳播 | Epoch：**2** <br> $F_1=0$ | Epoch：48 <br> $F_1=0.0025$ | Epoch：**2** <br> $F_2=0$ | Epoch：56 <br> $F_2=0.01$ | Epoch：**2** <br> $F_3=6\times10^{-33}$ | Epoch：13 <br> $F_3=5\times10^{-9}$ |

# 5. 結論與未來研究

為探索各種梯度下降演算法超參數的最佳值，本研究提出一個分析框架，分別分析梯度下降、動量梯度下降、自適應梯度、以及方均根傳播等方法在三個目標函數的平均誤差，並以最小化平均誤差來取得每個超參數的最佳值。並且，通過實驗證明，本研究提出的方法都可以在較少的 Epoch 數找出目標函數的最佳解。此外，在 3.5 節中討論各種梯度下降演算法超參數在不同目標函數的變化，歸納出幾個超參數設置原則，可以作為未來其他模型設置超參數時參考。

未來研究可以嘗試兩個方向：(1). 結合深度學習模型，修改目標函數，分析深度學習模型下各種梯度下降演算法超參數的最佳值、(2). 將本研究提出的分析

框架應用到群智能最佳化演算法的超參數最佳化，例如：粒子群最佳化(Particle Swarm Optimization, PSO)演算法的加速常數最佳化和關聯的論證。

# 參考文獻